\documentclass[runningheads]{llncs}
\usepackage[T1]{fontenc}
\usepackage{graphicx}
\usepackage{booktabs}
\usepackage{multirow}
\usepackage{amsmath}
\begin{document}
\title{Label-Free Foundational Model Selection for Medical Image Classification under Distribution Shift via Pseudo Label Discrepancy}
\titlerunning{Label-Free Model Selection under Distribution Shift}
\author{Juan Iñaki Larrea$^1$, Lucas Mansilla$^2$, Enzo Ferrante$^1$}
\authorrunning{Juan Iñaki Larrea, Lucas Mansilla, Enzo Ferrante}
%
\institute{$^1$Applied Artificial Intelligence Lab (LIAA), Computer Sciences Institute (ICC), Universidad de Buenos Aires - CONICET, Argentina\\
$^2$Research Institute for Signals, Systems and Computational Intelligence, sinc(i), Universidad Nacional del Litoral - CONICET, Argentina}
\maketitle
\begin{abstract}
Foundation models are increasingly deployed for medical image analysis. However, under the inter-institutional distribution shift typical of deployment, their performance varies widely and cannot be known without target-domain labels, which are rarely available. This leaves a practical question unresolved: given several candidate foundational models and labeled-data from a source domain, which one to deploy in an unlabeled target domain? We propose a label-free selection criterion built on SUDO, a framework for evaluating clinical AI systems without ground-truth annotations. SUDO partitions the unlabeled target data by predicted probability and, for each region, measures a pseudo-label discrepancy reflecting class contamination; aggregated across regions, this yields a score (AURCC) requiring neither target annotation nor fine-tuning. We show that AURCC can be used to rank a variety of vision-language models (BioMedCLIP, CXR-CLIP, CheXzero, MedCLIP, MedImageInsight, CLIP) on chest X-ray classification across three inter-hospital shift scenarios, under zero-shot and MLP-probe regimes. The AURCC ranking recovers the ground-truth ranking with Spearman $\rho$ up to $0.943$ ($p<0.05$). Against the natural baseline of ranking by held-out source accuracy, AURCC is competitive when the labeled source is large and yields a more accurate ranking once it is small; the regime of interest in resource-constrained settings.


\end{abstract}
\section{Introduction}

Machine learning has transformed medical image analysis, in some cases matching or exceeding specialist performance~\cite{litjens2017survey}. However, most supervised methods assume that training and deployment data are drawn from the same distribution, an assumption that rarely holds in practice: clinical images may come from different scanners, hospitals, patient populations, or acquisition protocols. This phenomenon, known as \emph{distribution shift}~\cite{quinonero2009dataset}, can degrade performance substantially~\cite{albadawy2018deep,zech2018variable}. Quantifying this degradation conventionally requires ground-truth annotations in the target domain, since standard metrics compare predictions against reference labels. In deployment, however, such annotations are typically unavailable: obtaining them requires expert radiologist time, which is costly and often inaccessible. This raises the question we address in this work: given a labeled source dataset and an unlabeled target domain, how can we rank candidate models by their expected target performance without annotations?

Foundation models have reshaped this landscape. Pre-trained on large image--text corpora, vision-language models such as CLIP and its biomedical variants produce general-purpose embeddings that transfer to downstream tasks with little or no supervision, via zero-shot (ZS) prompting, lightweight linear probe (LP) on frozen features or even training a shallow multi-layer perceptron with them (MP) ~\cite{radford2021clip,zhang2023biomedclip}. This is especially attractive in resource-constrained settings: the models are public and free, adaptation is cheap, and analysis can operate on embeddings orders of magnitude smaller than the images, cutting the communication cost of distributed learning and enabling data sharing through latent representations rather than raw scans~\cite{restrepo2024multimodal,fm_fed_embeddings,histo_embeddings}. Yet this generality is double-edged: many chest X-ray foundation models now exist, and their performance under inter-institutional shift varies widely and cannot be known \emph{a priori} without target labels. Choosing which to deploy is thus a key practical bottleneck, precisely the label-free ranking problem we address here.

\begin{figure}[t]
\centering
\includegraphics[width=\textwidth]{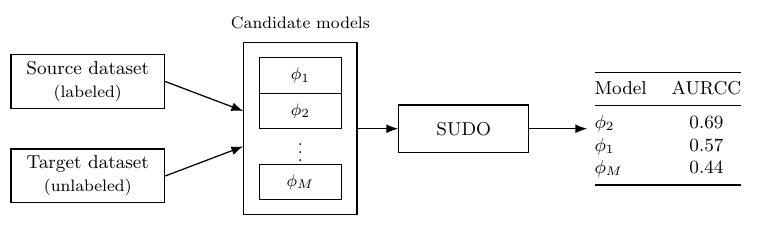}
\caption{Overview of the label-free model ranking framework. Given a labeled source dataset and an unlabeled target dataset, each candidate model $\phi_1, \ldots, \phi_M$ produces embeddings that drive the SUDO procedure, yielding a ranking of models by their AURCC without target labels.}
\label{fig:overview}
\end{figure}

We build on \textsc{SUDO}~\cite{kiyasseh2024sudo}, a framework that estimates the reliability of a deployed classifier without ground-truth labels by partitioning the target data according to the model's output probabilities and computing, for each region, a pseudo-label discrepancy index that reflects the degree of class contamination in that region. Aggregated across regions via a reliability--completeness curve, these discrepancies yield a single label-free score: the Area Under the Reliability-Completeness Curve (AURCC) that we reinterpret as a ranking signal across models. Both regimes we consider, ZS and MP, rely on source labels only; the target images are used unlabeled, only to generate the probabilities that drive the partitioning. A natural objection is that one could simply rank models by their accuracy on the held-out source data. We take this baseline and evaluate against it, showing that it is strong when the labeled source is large, but degrades when it is small. Interestingly, this is the situation faced by institutions with scarce annotation budgets. Fig.~\ref{fig:overview} gives an overview of the proposed label-free model ranking framework.

\noindent \textbf{Previous work.} Distribution shift is more often mitigated than measured: unsupervised domain adaptation aligns source and target representations~\cite{guan2021domain}, and test-time adaptation adjusts the model using only incoming target data~\cite{wang2021tent}. These approaches modify the model rather than assess it, leaving open the question we address: how to evaluate target performance without labels, and hence how to choose among candidate models. Estimating performance on unlabeled data remains an open problem in itself.  

Recently, Kiyasseh et al.~\cite{kiyasseh2024sudo} propose \textsc{SUDO}, a framework for evaluating clinical AI systems without ground-truth annotations. SUDO partitions the unlabeled target data into regions according to the model's predicted probabilities and, for each region, measures the discrepancy derived from auxiliary classifiers trained under opposing pseudo-label assumptions. Aggregating these discrepancies across regions yields a single label-free score, the AURCC. \textsc{SUDO} is framed as a tool for assessing a single deployed system, and the authors note that the AURCC can also serve to compare models. We take this observation and study it systematically in the setting of foundation model selection for medical imaging under distribution shift.
We benchmark six chest X-ray foundation models across three inter-hospital shift scenarios and two regimes of source-label usage, showing that the label-free ranking recovers ground-truth target performance (Spearman $\rho$ up to $0.943$, $p<0.05$) consistently.

\section{Methodology}
\label{sec:problem}

Let $\{\phi_1, \ldots, \phi_M\}$ denote a set of pre-trained foundation models. Given a labeled source dataset $\mathcal{D}_s = \{(x_i^s, y_i^s)\}_{i=1}^{N_s}$ and an unlabeled target dataset $\mathcal{D}_t = \{x_j^t\}_{j=1}^{N_t}$ drawn from a shifted distribution, our goal is to assign each model $\phi_m$ ($m=1,\ldots,M$) a label-free score $\text{AURCC}(\phi_m)$, computed without access to target labels $y^t$, such that ranking models by this score approximates the ranking induced by their true classification performance on $\mathcal{D}_t$. We evaluate this approximation against the ground-truth reference $\text{AUC}_{\text{target}}(\phi_m)$, the true classification AUC of $\phi_m$ on the full target set at its natural prevalence, computed post-hoc. We also compare against a natural baseline: ranking models by $\text{AUC}_{\text{source}}(\phi_m)$ on a held-out portion of the labeled source data.

\subsection{SUDO as a ranking criterion}
\label{sec:sudoprob}
Our ranking proxy builds on SUDO~\cite{kiyasseh2024sudo}. For a given model $\phi_m$, we compute its positive-class probabilities on the target samples and discretize them into $K$ quantile bins. Each bin $b$ contains $n=\frac{N_t}{K}$ \textbf{unlabeled target samples}, to which we assign a temporary pseudo-label and combine with an equal number of \textbf{ground-truth-labeled samples from the source-train split}, forming a binary training set. Two such sets are built under opposing assumptions: one pseudo-labels the bin as negative and adds source positives, the other pseudo-labels it as positive and adds source negatives. An auxiliary logistic regression classifier is trained on each set and evaluated on the held-out source split, yielding $s_0(b)$ and $s_1(b)$, the held-out AUC of the classifier under the negative and positive pseudo-label assumption, respectively. Then, the SUDO score for bin $b$ is
\begin{equation}
\text{SUDO}_b(\phi_m) = s_0(b) - s_1(b).
\end{equation}
A large $|\text{SUDO}_b(\phi_m)|$ indicates the bin is dominated by one class; values near zero indicate class contamination. The procedure is repeated over 300 stochastic runs and averaged.

Following Kiyasseh et al.~\cite{kiyasseh2024sudo}, the $K$ bin scores are aggregated into the Area Under the Reliability--Completeness Curve: sweeping symmetric quantile thresholds from the extremes of the probability distribution toward the center traces reliability (mean $|\text{SUDO}|$ over the selected bins) against completeness (fraction of target samples covered), and the trapezoidal area under this curve, $\text{AURCC}(\phi_m) \in [0,1]$, is our label-free ranking score.

As an experimental control, the target set is subsampled to match the source positive rate. This ensures comparable bin composition across scenarios but is not a requirement of the method. In the MLP-Probe scenario, where we train a shallow MLP on top of the foundational embeddings (see next section for more details), auxiliary classifiers operate on 128-dimensional projected features of the probe; in ZS, on the raw $D$-dimensional embeddings produced by the image encoder $\phi_m$.

\subsection{Zero-shot and MLP-probe variants}
\label{sec:regimes}
We evaluate two strategies for obtaining the target probabilities $p(x^t)$ that drive the SUDO partitioning.

\textbf{Zero-Shot (ZS).} No task-specific training is performed. Following standard CLIP-style zero-shot classification, we construct an ensemble of 10 positive and 10 negative text prompts per class, encode and mean-pool each prompt set into class prototypes $t_{\text{pos}}, t_{\text{neg}}$, and compute $p(x) = \text{softmax}\big([x \cdot t_{\text{pos}},\ x \cdot t_{\text{neg}}]\big)_0$, with all embeddings L2-normalized.

\textbf{MLP Probe (MP).} With the pre-trained image encoder kept frozen, we train a lightweight classification head on top of its embeddings: a fully connected layer ($D \to 128$), ReLU, dropout ($p=0.3$), and a final fully connected layer ($128 \to 1$). This shallow MLP head, denoted $\text{probe}(\cdot)$, is trained only on source-train embeddings and labels (Adam optimizer, lr $10^{-3}$, weight decay $10^{-4}$, batch size 256, 20 epochs with early stopping). At inference, the target probability is $p(x) = \sigma(\text{probe}(x))$.

\subsection{Data, models, and compute resources}
\label{sec:setup}

We evaluate six chest X-ray foundation models: BioMedCLIP \cite{zhang2023biomedclip}, CheXzero \cite{tiu2022chexzero}, CXR-CLIP \cite{you2023cxrclip}, MedCLIP \cite{wang2022medclip}, MedImageInsight \cite{codella2024medimageinsight}, and CLIP-base \cite{radford2021clip} (a general-domain baseline). All models produce zero-shot class probabilities from image--text similarity. The task is binary pneumonia classification, shared across all datasets. We use three source datasets that appear in the pre-training corpora of the medical models (MIMIC-CXR \cite{johnson2019mimic}, NIH ChestX-ray14 \cite{wang2017chestxray8}, CheXpert \cite{irvin2019chexpert}) and PadChest \cite{bustos2020padchest} as the target domain, chosen because it is absent from every model's pre-training data and differs from all three sources in institution, making the resulting pairs realistic inter-hospital shift scenarios.

Each dataset is filtered to the samples carrying a binary pneumonia label. For CheXpert this excludes uncertain and unmentioned labels, leaving a small labeled subset ($N=563$). NIH is extremely imbalanced ($1.3\%$ positive), which would leave most probability bins with no positive samples; we therefore subsample the negatives to reach a $30\%$ positive rate. Each source is then split $80/20$ into a training portion, used to fit the MP head and to draw the labeled samples for pseudo-labeling, and a held-out portion, used both to evaluate the SUDO auxiliary classifiers and to compute the source-AUC baseline. Table~\ref{tab:splits} reports sizes and positive rates.

\begin{table}[h]
\caption{Dataset sizes and positive rates after filtering for a binary pneumonia label. Each source is split 80/20 into train and held-out portions.}\label{tab:splits}
\centering
\begin{tabular}{lrc}
\toprule
Dataset & $N$ & Positive rate \\
\midrule
MIMIC-CXR & 37{,}457 & 41.6\% \\
NIH ChestX-ray14 & 4{,}770 & 30.0\% \\
CheXpert & 563 & 67.5\% \\
PadChest & 88{,}018 & 3.2\% \\
\bottomrule
\end{tabular}
\end{table}

Embedding extraction was performed on a single NVIDIA Titan Xp (12GB), with batch size adjusted per model to fit memory constraints. SUDO evaluation, comprising auxiliary classifier training, MP training and AURCC computation, is CPU-only and uses scikit-learn logistic regression.

\section{Results}
\subsection{Label-free model ranking}

Table~\ref{tab:results} reports the Spearman correlation between the ranking induced by $\text{AURCC}$ and the ground-truth $\text{AUC}_{\text{target}}$, across $K \in \{5,10,20\}$ and both regimes (ZS and MP). For comparison, we also report the Spearman~$\rho$ obtained by ranking models according to their $\text{AUC}_{\text{source}}$ on the held-out source split, the natural baseline when only source labels are available.

\begin{figure}[h]
    \centering
    \includegraphics[width=\textwidth]{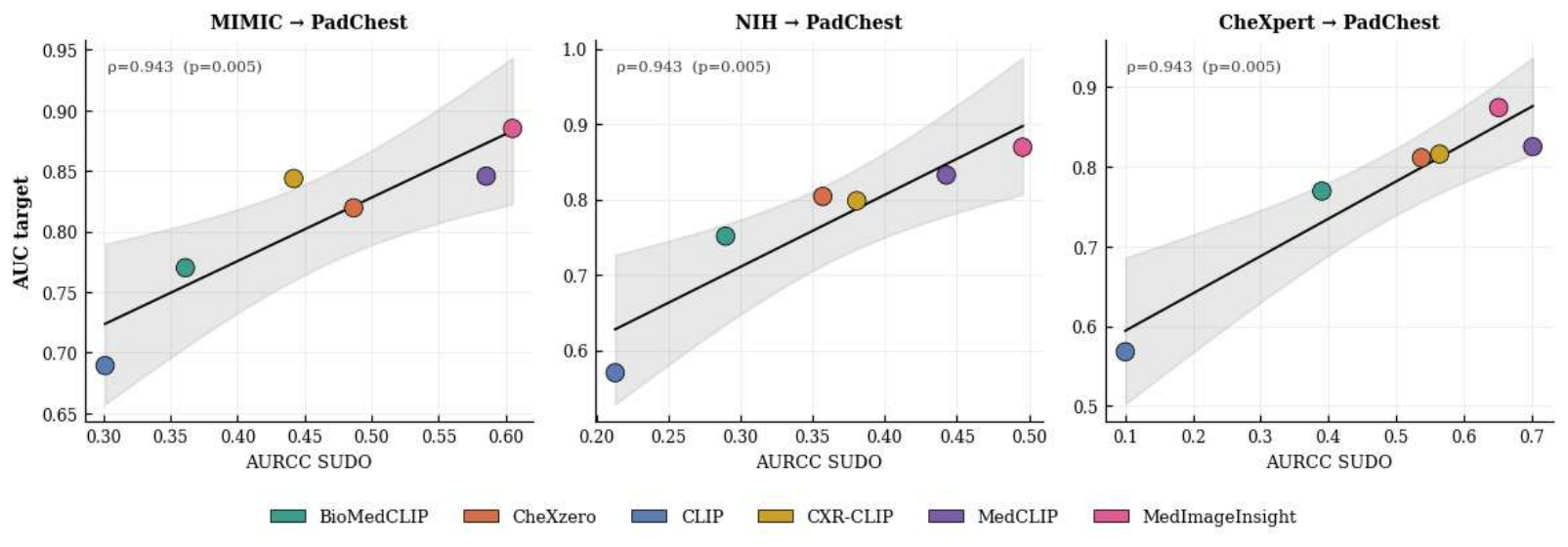}
    \caption{Label-free $\text{AURCC}$ (x-axis, MLP-probe, $K=10$) versus true $\text{AUC}_{\text{target}}$ (y-axis) for the three source--target scenarios. Each point is one of the six foundation models. The shaded band indicates the 95\% confidence interval of the regression line.}
    \label{fig:scatter}
\end{figure}

In the MP regime, the AURCC ranking reaches $\rho = 0.943$ ($p<0.05$) in every scenario and every value of $K$, showing that the ranking is stable and does not depend on the number of regions. The source-AUC baseline achieves $\rho = 1.000$ for MIMIC ($N_{\text{source-heldout}}=7{,}506$) and $\rho = 0.943$ for CheXpert ($N_{\text{source-heldout}}=113$), but drops to $\rho = 0.829$ for NIH ($N_{\text{source-heldout}}=954$), where AURCC provides a strictly better ranking. In the ZS regime, correlations remain moderate to high ($\rho$ from $0.714$ to $0.943$), though the source-AUC baseline is stronger in the MIMIC and NIH scenarios, while AURCC is stronger in CheXpert. Fig.~\ref{fig:scatter} shows the MP results: the label-free score orders the models consistently with their true target AUC, with the general-domain baseline correctly placed at the bottom in all three scenarios.

\begin{table}[h]
\caption{Spearman $\rho$ between model rankings and ground-truth $\text{AUC}_{\text{target}}$, for $M=6$ foundation models. AURCC = label-free SUDO score; $\text{AUC}_{\text{source}}$ = held-out source baseline. $^*p<0.05$.}\label{tab:results}
\centering
\begin{tabular}{llcccc}
\toprule
 & & \multicolumn{3}{c}{AURCC vs.\ AUC$_{\text{target}}$} & AUC$_{\text{source}}$ vs.\ AUC$_{\text{target}}$ \\
\cmidrule(lr){3-5} \cmidrule(lr){6-6}
Source $\to$ Target & Regime & K=5 & K=10 & K=20 &  \\
\midrule
MIMIC $\to$ PadChest    & ZS & 0.886* & 0.714 & 0.714 & 0.943* \\
MIMIC $\to$ PadChest    & MP & 0.943* & 0.943* & 0.943* & 1.000* \\
NIH $\to$ PadChest      & ZS & 0.886* & 0.886* & 0.886* & 1.000* \\
NIH $\to$ PadChest      & MP & 0.943* & 0.943* & 0.943* & 0.829* \\
CheXpert $\to$ PadChest & ZS & 0.943* & 0.943* & 0.943* & 0.886* \\
CheXpert $\to$ PadChest & MP & 0.943* & 0.943* & 0.943* & 0.943* \\
\bottomrule
\end{tabular}
\end{table}

Fig.~\ref{fig:scatter} shows the ordering is not perfect among the intermediate models: in each scenario, a pair of models is locally swapped relative to their true target AUC. These swaps involve models whose target AUC is otherwise close, and do not affect the overall separation between the general-domain baseline and the medical models. This suggests that AURCC's advantage is tied to task-specific adaptation: once source labels shape a task-specific decision boundary, as in MP, AURCC can exploit structure in the target that a single source-accuracy estimate cannot. In ZS, with no such adaptation, neither criterion has task-specific signal to draw on beyond the model's general-purpose pre-training. The gap between the two criteria opens once labels are used to shape that boundary, which we examine next.

\subsection{Sensitivity to held-out source size}

The preceding analysis uses the held-out source for both the AURCC computation and the source-AUC baseline. However, in resource-constrained settings (RCS), the labeled source may be small. We therefore study how each ranking criterion degrades as the held-out source is subsampled, both to fixed fractions of the held-out set (75\%, 50\%, 25\%) and to fixed sample sizes ($N=100$, $N=50$). Subsampling affects both methods: the source-AUC baseline is recomputed on fewer samples, and the full SUDO pipeline is re-run with the subsampled source used to evaluate the auxiliary classifiers. The training portion used for MLP-probe fitting and as labeled source for pseudo-labeling remains fixed throughout. Each of the 20 repetitions recomputes both criteria from scratch on the resampled held-out source. Table~\ref{tab:sensitivity} reports the resulting mean $\rho$ at each size.

\begin{table}[h]
\caption{Mean Spearman $\rho$ (over 20 repetitions) of AURCC and source-AUC rankings versus $\text{AUC}_{\text{target}}$, as a function of source held-out size. MP regime, $K=10$.}\label{tab:sensitivity}
\centering
\begin{tabular}{llccccc}
\toprule
Source $\to$ Target & Method & 100\% & 50\% & 25\% & $N$=100 & $N$=50 \\
\midrule
\multirow{2}{*}{MIMIC $\to$ PadChest} & AURCC & 0.943 & 0.943 & 0.943 & 0.851 & \textbf{0.683} \\
 & AUC$_{\text{source}}$ & \textbf{1.000} & \textbf{0.997} & \textbf{0.989} & \textbf{0.916} & 0.605 \\
\midrule
\multirow{2}{*}{NIH $\to$ PadChest} & AURCC & \textbf{0.994} & \textbf{0.977} & \textbf{0.963} & \textbf{0.911} & \textbf{0.877} \\
 & AUC$_{\text{source}}$ & 0.829 & 0.809 & 0.734 & 0.749 & 0.752 \\
\midrule
\multirow{2}{*}{CheXpert $\to$ PadChest} & AURCC & \textbf{0.980} & \textbf{0.897} & --- & \textbf{0.977} & \textbf{0.880} \\
 & AUC$_{\text{source}}$ & 0.943 & 0.749 & --- & 0.897 & 0.706 \\
\bottomrule
\end{tabular}
\end{table}

\begin{figure}[t!]
    \centering
    \includegraphics[width=\textwidth]{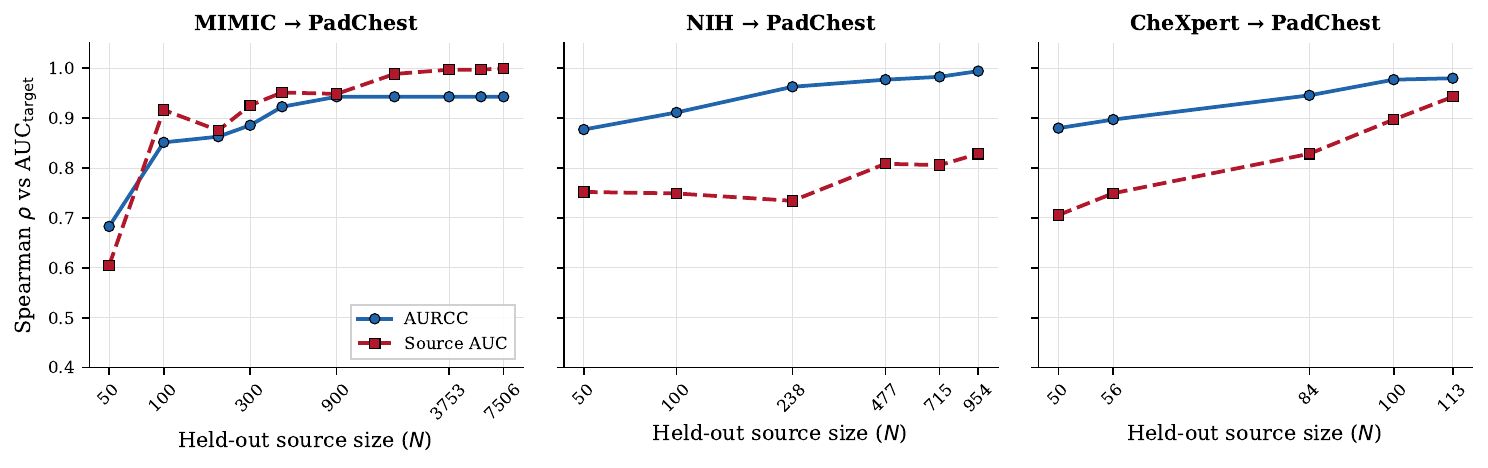}
    \caption{Sensitivity of AURCC and source-AUC rankings to held-out source size (MP, $K=10$). Each point is the mean Spearman $\rho$ over 20 stratified subsampling repetitions.}
    \label{fig:sensitivity}
\end{figure}

The results reveal a clear pattern (Table~\ref{tab:sensitivity}, Fig.~\ref{fig:sensitivity}). For MIMIC ($N_{\text{source-heldout}}=7{,}506$), the source-AUC baseline is near-perfect at full size ($\rho=1.000$) and AURCC does not exceed it until the source is reduced to $N=50$, where AURCC achieves $\rho=0.683$ versus $\rho=0.605$. However, for NIH ($N_{\text{source-heldout}}=954$) and CheXpert ($N_{\text{source-heldout}}=113$), AURCC consistently outperforms the source-AUC baseline at every size tested, with the gap widening as the source shrinks: for NIH at $N=100$, AURCC achieves $\rho=0.911$ versus $\rho=0.749$ for source AUC. At the most extreme reduction tested ($N=50$), AURCC outperforms the source-AUC baseline in all three scenarios: when the labeled sample is this small, drawing on the structure of the unlabeled target data yields a more reliable ranking than a point estimate computed on a handful of source examples.
CheXpert's 25\% fraction ($N=28$) exceeds the available source so is omitted.

\section{Discussion} In this work, we propose a label-free criterion for ranking foundation models under distribution shift. Across three inter-hospital scenarios toward PadChest, the AURCC ranking correlates strongly with ground-truth target performance. In MP, where the models have a trained classifier from the source dataset, the AURCC ranking is competitive with or superior to the natural source-AUC baseline. In ZS, where no classifier is trained on the source, the ranking relies solely on the geometry of pre-trained embeddings so the source-AUC baseline and AURCC perform comparably. The advantage of AURCC emerges precisely when it matters most: when the labeled source is small. For NIH ($N=954$) and CheXpert ($N=113$), AURCC outperforms source AUC at every fraction tested, and in all scenarios produces a better ranking for $N=50$. This suggests that AURCC captures structural information about the target distribution, via the pseudo-label discrepancy across probability regions, that goes beyond what a simple aggregate metric on source data can provide.
Our evaluation is restricted to a single binary task (pneumonia) and a single target domain (PadChest). While the three source datasets provide independent shift scenarios, generalization to other tasks, modalities, and multi-class settings remains to be validated.\\

\noindent \textbf{Impact in RCS.} 
In resource-constrained settings, foundation models are particularly useful, as they are public, free, and adaptation (zero-shot, linear probe, or MLP probe) is cheap. What the setting lacks is a way to tell, among the many chest X-ray foundation models now available, which to trust at a new institution. That conventionally requires target labels, and it is precisely in these settings where they are scarcest. Our criterion reduces this dependency, as it ranks candidates using unlabeled target images, easing the selection bottleneck. Because the whole pipeline consumes only embeddings and runs on CPU, an institution can rank models locally on commodity hardware. Our study also shows that the source-AUC baseline degrades as the sample shrinks (Table~\ref{tab:sensitivity}). Particularly in these scenarios, AURCC draws its signal from the unlabeled target and degrades far more gracefully, giving better ranking quality in small-source scenarios, more likely to be found in resource-constrained settings.



%
\subsubsection*{Disclosure of Interests.}
The authors have no competing interests to declare that are relevant to the content of this article.
\bibliographystyle{splncs04}
\bibliography{references}

\begin{thebibliography}{10}
\providecommand{\url}[1]{\texttt{#1}}
\providecommand{\urlprefix}{URL }
\providecommand{\doi}[1]{https://doi.org/#1}

\bibitem{albadawy2018deep}
Albadawy, E.A., Saha, A., Mazurowski, M.A.: Deep learning segmentation of brain tumors: impact of cross-institutional training and testing. Medical Physics  \textbf{45}(3),  1150--1158 (2018)

\bibitem{bustos2020padchest}
Bustos, A., Pertusa, A., Salinas, J.M., de~la Iglesia-Vay{\'a}, M.: Padchest: A large chest x-ray image dataset with multi-label annotated reports. Medical Image Analysis  \textbf{66},  101797 (2020). \doi{10.1016/j.media.2020.101797}

\bibitem{histo_embeddings}
Chelebian, E., Ciompi, F., W\"ahlby, C.: Seeded iterative clustering for histology region identification. arXiv preprint arXiv:2211.07425  (2022)

\bibitem{codella2024medimageinsight}
Codella, N.C.F., Jin, Y., Jain, S., Gu, Y., Lee, H.H., Ben~Abacha, A., Santamaria-Pang, A., Guyman, W., Sangani, N., Zhang, S., Poon, H., Hyland, S., Bannur, S., Alvarez-Valle, J., Li, X., Garrett, J., McMillan, A., Rajguru, G., Maddi, M., Vijayrania, N., Bhimai, R., Mecklenburg, N., Jain, R., Holstein, D., Gaur, N., Aski, V., Hwang, J.N., Lin, T., Tarapov, I., Lungren, M., Wei, M.: Medimageinsight: An open-source embedding model for general domain medical imaging. arXiv preprint arXiv:2410.06542  (2024)

\bibitem{guan2021domain}
Guan, H., Liu, M.: Domain adaptation for medical image analysis: a survey. IEEE Transactions on Biomedical Engineering  \textbf{69}(3),  1173--1185 (2021)

\bibitem{irvin2019chexpert}
Irvin, J., Rajpurkar, P., Ko, M., Yu, Y., Ciurea-Ilcus, S., Chute, C., Marklund, H., Haghgoo, B., Ball, R., Shpanskaya, K., Seekins, J., Mong, D.A., Halabi, S.S., Sandberg, J.K., Jones, R., Larson, D.B., Langlotz, C.P., Patel, B.N., Lungren, M.P., Ng, A.Y.: Chexpert: A large chest radiograph dataset with uncertainty labels and expert comparison. In: Proceedings of the AAAI Conference on Artificial Intelligence. vol.~33, pp. 590--597 (2019). \doi{10.1609/aaai.v33i01.3301590}

\bibitem{johnson2019mimic}
Johnson, A.E.W., Pollard, T.J., Berkowitz, S.J., Greenbaum, N.R., Lungren, M.P., Deng, C.y., Mark, R.G., Horng, S.: Mimic-cxr, a de-identified publicly available database of chest radiographs with free-text reports. Scientific Data  \textbf{6}(1), ~317 (2019). \doi{10.1038/s41597-019-0322-0}

\bibitem{kiyasseh2024sudo}
Kiyasseh, D., Cohen, A., Jiang, C., Altieri, N.: A framework for evaluating clinical artificial intelligence systems without ground-truth annotations. Nature Communications  \textbf{15}(1), ~1808 (2024)

\bibitem{litjens2017survey}
Litjens, G., Kooi, T., Bejnordi, B.E., Setio, A.A.A., Ciompi, F., Ghafoorian, M., Van Der~Laak, J.A., Van~Ginneken, B., S{\'a}nchez, C.I.: A survey on deep learning in medical image analysis. Medical Image Analysis  \textbf{42},  60--88 (2017)

\bibitem{fm_fed_embeddings}
Lohmann, J.J.G., Witte, A., Maier, A., Saak, C.C., Sauter, G., Zimmermann, M., Bonn, S., Baumbach, J.: On the power and limits of foundation model image embeddings for privacy-preserving federated learning. Array  \textbf{29},  100725 (2026)

\bibitem{quinonero2009dataset}
Qui{\~n}onero-Candela, J., Sugiyama, M., Schwaighofer, A., Lawrence, N.D.: Dataset Shift in Machine Learning. MIT Press (2009)

\bibitem{radford2021clip}
Radford, A., Kim, J.W., Hallacy, C., Ramesh, A., Goh, G., Agarwal, S., Sastry, G., Askell, A., Mishkin, P., Clark, J., Krueger, G., Sutskever, I.: Learning transferable visual models from natural language supervision. In: Proceedings of the 38th International Conference on Machine Learning (ICML). Proceedings of Machine Learning Research, vol.~139, pp. 8748--8763. PMLR (2021)

\bibitem{restrepo2024multimodal}
Restrepo, D., Wu, C., Cajas, S.A., Nakayama, L.F., Celi, L.A., L{\'o}pez, D.M.: Multimodal deep learning for low-resource settings: a vector embedding alignment approach for healthcare applications. arXiv preprint arXiv:2406.02601  (2024)

\bibitem{tiu2022chexzero}
Tiu, E., Talius, E., Patel, P., Langlotz, C.P., Ng, A.Y., Rajpurkar, P.: Expert-level detection of pathologies from unannotated chest x-ray images via self-supervised learning. Nature Biomedical Engineering  \textbf{6}(12),  1399--1406 (2022). \doi{10.1038/s41551-022-00936-9}

\bibitem{wang2021tent}
Wang, D., Shelhamer, E., Liu, S., Olshausen, B., Darrell, T.: Tent: Fully test-time adaptation by entropy minimization. In: International Conference on Learning Representations (ICLR) (2021)

\bibitem{wang2017chestxray8}
Wang, X., Peng, Y., Lu, L., Lu, Z., Bagheri, M., Summers, R.M.: Chestx-ray8: Hospital-scale chest x-ray database and benchmarks on weakly-supervised classification and localization of common thorax diseases. In: Proceedings of the IEEE Conference on Computer Vision and Pattern Recognition (CVPR). pp. 2097--2106 (2017)

\bibitem{wang2022medclip}
Wang, Z., Wu, Z., Agarwal, D., Sun, J.: Medclip: Contrastive learning from unpaired medical images and text. In: Proceedings of the 2022 Conference on Empirical Methods in Natural Language Processing (EMNLP). pp. 3876--3887. Association for Computational Linguistics (2022). \doi{10.18653/v1/2022.emnlp-main.256}

\bibitem{you2023cxrclip}
You, K., Gu, J., Ham, J., Park, B., Kim, J., Hong, E.K., Baek, W., Roh, B.: Cxr-clip: Toward large scale chest x-ray language-image pre-training. In: Medical Image Computing and Computer Assisted Intervention (MICCAI). Lecture Notes in Computer Science, vol. 14221, pp. 101--111. Springer (2023). \doi{10.1007/978-3-031-43895-0_10}

\bibitem{zech2018variable}
Zech, J.R., Badgeley, M.A., Liu, M., Costa, A.B., Titano, J.J., Oermann, E.K.: Variable generalization performance of a deep learning model to detect pneumonia in chest radiographs: a cross-sectional study. PLoS Medicine  \textbf{15}(11),  e1002683 (2018)

\bibitem{zhang2023biomedclip}
Zhang, S., Xu, Y., Usuyama, N., Xu, H., Bagga, J., Tinn, R., Preston, S., Rao, R., Wei, M., Valluri, N., Wong, C., Tupini, A., Wang, Y., Mazzola, M., Shukla, S., Liden, L., Gao, J., Crabtree, A., Piening, B., Bifulco, C., Lungren, M.P., Naumann, T., Wang, S., Poon, H.: Biomedclip: a multimodal biomedical foundation model pretrained from fifteen million scientific image-text pairs. arXiv preprint arXiv:2303.00915  (2023)

\end{thebibliography}
\end{document}